\documentclass[sigconf,natbib=true]{acmart}

\usepackage{dsfont}
\usepackage{array}
\usepackage{makecell} 
\usepackage{CJKutf8}

\usepackage{booktabs}
\usepackage{tabularx}
\usepackage{multirow}
\usepackage{graphicx}

\AtBeginDocument{%
	\providecommand\BibTeX{{%
			\normalfont B\kern-0.5em{\scshape i\kern-0.25em b}\kern-0.8em\TeX}}}

\newcommand{\secref}[1]{Section \ref{#1}}
\newcommand{\figref}[1]{Figure \ref{#1}}

\newcommand{\tableref}[1]{Table \ref{#1}}

\newcommand{\cut}[1]{}

\begin{document}
\begin{CJK*}{UTF8}{gbsn}
\title{A Dataset of Open-Domain Question Answering with Multiple-Span Answers}  

\author{Zhiyi Luo}
\email{luozhiyi@zstu.edu.cn}
\affiliation{%
	\institution{Zhejiang Sci-Tech University}
	\streetaddress{No. 928, No. 2 street, Baiyang street}
	\city{Hangzhou}
	\state{Zhejiang}
	\country{China}
	\postcode{310018}
}
\orcid{0002-2206-1926}

\author{Yingying Zhang}
\email{272831920@qq.com}
\affiliation{%
	\institution{Zhejiang Sci-Tech University}
	\streetaddress{No. 928, No. 2 street, Baiyang street}
	\city{Hangzhou}
	\state{Zhejiang}
	\country{China}
	\postcode{43017-6221}
}

\author{Shunyun Luo}
\authornotemark[1]
\email{shuyunluo@zstu.edu.cn}
\affiliation{%
  \institution{Zhejiang Sci-Tech University}
  \streetaddress{No. 928, No. 2 street, Baiyang street}
  \city{Hangzhou}
  \state{Zhejiang}
  \country{China}
  \postcode{310018}
}

\author{Ying Zhao}
\email{308956149@qq.com}
\affiliation{%
	\institution{Zhejiang Sci-Tech University}
	\streetaddress{No. 928, No. 2 street, Baiyang street}
	\city{Hangzhou}
	\state{Zhejiang}
	\country{China}
	\postcode{310018}
}

\author{Wentao Lyu}
\email{alvinlwt@zstu.edu.cn}
\affiliation{%
	\institution{Zhejiang Sci-Tech University}
	\streetaddress{No. 928, No. 2 street, Baiyang street}
	\city{Hangzhou}
	\state{Zhejiang}
	\country{China}
	\postcode{310018}
}

%

\renewcommand{\shortauthors}{Trovato et al.}

\begin{abstract}
 Multi-span answer extraction, also known as the task of multi-span question answering (MSQA), is critical for real-world applications, as it requires extracting multiple pieces of information from a text to answer complex questions. Despite the active studies and rapid progress in English MSQA research, there is a notable lack of publicly available MSQA benchmark in Chinese. Previous  efforts for constructing  MSQA datasets predominantly emphasized entity-centric contextualization, resulting in a bias towards collecting factoid questions and potentially overlooking questions requiring more detailed descriptive responses. To overcome these limitations, we present CLEAN, a comprehensive Chinese multi-span question answering dataset that involves a wide range of open-domain subjects with a substantial number of  instances requiring descriptive answers. Additionally, we provide established models from relevant literature as baselines for CLEAN. Experimental results and analysis show the characteristics and challenge of the  newly proposed CLEAN dataset for the community. Our dataset, CLEAN, will be publicly released at~\url{zhiyiluo.site/misc/clean_v1.0_sample.json}.
\end{abstract}




\maketitle

\section{Introduction}
\label{sec:introduction}

Extractive question answering, commonly known as reading comprehension (RC), which aims to answer a user's question by finding short text segments (i.e., {\em answer spans}) from the given context, has been actively studied and achieved rapid progress in recent years. 
Earlier studies~\cite{rajpurkar2016squad,rajpurkar2018know,yang2018hotpotqa,zaheer2020big} restrict the answer extracted from the context to a single text span.
However, a comprehensive answer to a real-world question could consist of a series of non-contiguous spans, or even the question itself could have multiple intents, 
where the answer to each intent is composed of one or more spans drawn from the input.

Recent research efforts have explored various  datasets~\cite{DBLP:conf/emnlp/DasigiLMSG19,DBLP:journals/tacl/ClarkPNCGCK20,DBLP:journals/tacl/KwiatkowskiPRCP19,DBLP:journals/ipm/MalhasE22,li2022multispanqa}  and      models~\cite{lee2023liquid,DBLP:journals/ipm/Liu0GC23,DBLP:conf/cikm/HuangZN023,DBLP:conf/acl/ZhangL0LF023} for open-domain multi-span question answering (MSQA).
However, most existing MSQA datasets are in English~\cite{DBLP:conf/naacl/DuaWDSS019,DBLP:conf/emnlp/DasigiLMSG19,DBLP:journals/tacl/KwiatkowskiPRCP19,li2022multispanqa}, and there is a noticeable scarcity of publicly available Chinese MSQA datasets. Although CMQA~\cite{DBLP:conf/coling/JuWZZ0022} recently stands out as the first public multi-span extraction QA dataset in Chinese, it focuses on specialized domains such as  healthcare and education, rather than the open domain, and formulates a new task of conditional question answering, which may not directly address the requirements of traditional question answering scenarios. 

\begin{table*}[!h]
	\small
	\caption{Examples of questions and answers.}
	\begin{tabular}{c|c|c} 
		\hline
		\addlinespace
		\vspace{-1ex}
		\textbf{Dataset}   & \textbf{MultiSpanQA}   &  \textbf{CLEAN} \\
		\addlinespace
		\hline
		Question  &  Who wrote the song if you could see me now?  & {\makecell[l]{
				被蚊子咬后有哪些比较好的消肿方式？
				\\ (What are some effective ways to reduce swelling after being bitten by mosquitoes?)}} \\ \hline
		Context  & {\makecell[l]{
				``If You Could See Me Now'' is a 1946 jazz \\standard,
				composed by {\color{blue}Tadd Dameron}. He \\wrote it especially for vocalist 
				Sarah Vaughan, \\a frequent collaborator. Lyrics were written\\ by {\color{blue}Carl Sigman} 
				and it became one of her \\signature songs, inducted into the 
				Grammy \\Hall of Fame in 1998.}}
		&  {\makecell[l]{...方法有很多种。一种是利用{\color{blue}大蒜片敷患处}，可明显缓解疼痛、止痒...另一种\\方法是{\color{blue}使用中成药}，如南通蛇药片、新癀片等，...
				此外，{\color{blue}西瓜皮也可涂擦患处}，\\有利于止痒、消肿...\\
				(...there are many methods available. One method is to apply {\color{blue}garlic slices to the} \\{\color{blue}affected area}, which can significantly alleviate pain and itchiness... Another method \\is to {\color{blue}use Chinese patent}{\color{blue}medicines} such as Nantong Sheyao Pian, Xinhuang Pian, etc.\\
				... Additionally, {\color{blue}rubbing watermelon rind on the affected area} can be beneficial for \\relieving itchiness and reducing swelling...) }}
		\\ \hline
		Answer &{\makecell[l]{Segment1: Tadd Dameron\\
				Segment2: Carl Sigman}}  &
		{\makecell[l]{Segment1:大蒜片敷患处(garlic slices to the affected area)\\
				Segment2:使用中成药(use Chinese patent medicines)\\
				Segment3:西瓜皮也可涂擦患(rubbing watermelon rind on the affected area)}}
		\\ \hline
	\end{tabular}
	\label{tab:intro}
\end{table*}

Furthermore, the current methodology for constructing open-domain MSQA datasets  is crafted to extract answers from organized, entity-centric  contexts, typically sourced from Wikipedia pages. This approach is inclined towards collecting factoid questions, where the expected answers are \emph{entities}, such as  \emph{Person} and \emph{Location}. However, it is less effective at capturing questions that require longer descriptive answers which are essential in many scenarios. 
For example, the popular MultiSpanQA~\cite{li2022multispanqa} dataset favors questions with short span answers, averaging 2.3 tokens in length, where  approximately 76.3\% of the answers are of the entity type, only 16.4\% are descriptive, and the  rest are numeric.
This  methodology struggles to contextualize many of these inquiries due to a semantic gap between user intentions and the entity-centric nature of the context.

To address the scarcity of  Chinese multi-answer RC datasets and to bridge the observed semantic gap, we propose a new \textbf{C}hinese mu\textbf{L}ti-span qu\textbf{E}stion \textbf{AN}swering (CLEAN) dataset, which features both multi-answer and single-answer examples from an open domain and supports to cast RC as an answer extraction task.  CLEAN is distinctively constructed by extracting question-context pairs from a large-scale knowledge Q\&A sharing platform (i.e., BaiduZhidao\footnote{https://zhidao.baidu.com/}), rather than relying on Wikipedia for context retrieval. 
Moreover, the context in CLEAN is carefully curated from long answers provided by respondents, effectively bridging the semantic gap. This approach ensures that the question intents are appropriately addressed and overcomes the limitations of previous datasets in terms of question selection. 
\tableref{tab:intro} (right) presents a specific example from the CLEAN dataset, where the question is obtained from BaiduZhidao, and the context is selected from the original long answers associated with that question. 
Note that CLEAN includes a high proportion of descriptive answers, approximately 76\%, offering a more comprehensive and nuanced resource for further research into Chinese RC tasks.

Our contributions are twofold: first, we create a new Chinese multi-answer RC dataset named CLEAN, which covers a wide range of open-domain question topics and supports to cast RC as an answer extraction task. Second, we establish models drawn from related literature as the baselines of CLEAN. Experiment results showcase the challenge of extracting longer descriptive answers, which constitute the majority in the  CLEAN dataset.
\section{Dataset Construction}
\label{sec:dataset}
In this section, we describe the process of constructing the CLEAN dataset and then show the overview statistics of the newly created dataset.

\subsection{Data Collection}
\label{sec:collection}
Samples in CLEAN are collected from a large-scale Chinese online knowledge Q\&A sharing platform which is full of open-domain questions with crafted long answers from public users. These questions, which have been contributed over two decades through crowd-sourcing efforts, cover a wide array of subjects.
To pursue a broad coverage of topics for CLEAN, we randomly crawled one million questions across 29 popular subjects in the open domain, including people (\emph{人物}), celestial bodies (\emph{天体}), flora and fauna (\emph{动植物}), landmarks (\emph{景点景观}), among others, each accompanied by well-thought-out long answers. 

Next, we employ two key strategies to increase the proportion of multi-answer instances relative to single-answer instances: 1) selectively choosing questions that contain  keywords such as \emph{如何/怎样}(e.g. \emph{how}) or \emph{哪些} (e.g.\emph{ what/which ones}), as these types of questions typically require multipart descriptions; and 2) translating all the crawled questions into English and specifically targeting questions featuring plural nouns. 
Note that, a question may have a number of long answers created by different users, and we only consider those with more than 3 likes as potential context candidates.
For context candidates that exceed the token limit (i.e. 350 tokens), we split them into chunks of up to 350 tokens based on BERT's tokenizer, while preserving sentence boundaries, and then pair each chunk with the  corresponding question to create one sample.
Then, we get about 32,000 filtered samples and build our dataset from these samples, following the annotation principles detailed in~\secref{sec:annotation}. 

\subsection{Annotation Process}
\label{sec:annotation}
The annotation process is realized by three full-time annotators using the Brat\footnote{https://zhidao.baidu.com/} annotation tool~\cite{stenetorp2012brat}. 
Our annotators are educated to pick up high-quality contexts from the candidate answers for each question and identify one or more answer spans within the context that can effectively answer the given question. 
During this phase, we discard samples that contain offensive content, as well as those with incomplete questions, unclear intents, or answers that cannot be extracted from the associated context. To further ensure the reliability of the annotations, we calculate the Fleiss' Kappa score to measure the inter-annotator agreement among different annotators for each sample. Only samples achieving a Fleiss' Kappa score greater than 0.5 are retained. 
Finally, we have curated a diverse collection of 9,063 samples, including both multi-answer and single-answer instances. Note that our CLEAN dataset demonstrates a high level of inter-annotator agreement, with an average Fleiss' Kappa score of 0.739, indicating that the annotations are consistent and reliable.

\subsection{Dataset Statistics}
\label{sec:statistics}
In previous datasets such as DROP~\cite{DBLP:conf/naacl/DuaWDSS019} and Quoref~\cite{DBLP:conf/emnlp/DasigiLMSG19}, the proportion of multi-answer instances is relatively low, at around 6\% and 10\%, respectively. 
More recently, the MultiSpanQA  dataset~\cite{li2022multispanqa} has been proposed 100\% for English multi-answer RC. The CLEAN dataset consists of 9,063 samples in total, including over 4.2K multi-answer instances, which accounts for about 46\%. 
~\tableref{tab:statistics} shows the overview statistics comparison, where the \textit{answer amount} refers to average number of answer span per sample and \textit{answer length} refers to the average length per answer span.
Note that only multi-answer samples are compared.
\begin{table}[!h]
	\small
	\caption{Overview statistics comparison of CLEAN with other multi-answer RC datasets.}
	\label{tab:statistics}
	\begin{tabular}{l@{\hspace{2.2pt}}c@{\hspace{5pt}}c@{\hspace{6pt}}ccc}
		\toprule
		\textbf{Dataset} 
		& \textbf{LANG}  & \textbf{\#QA} & \makecell{\textbf{Question} \\ \textbf{Source}} & \makecell{\textbf{Answer} \\ \textbf{Amount}} &\makecell{\textbf{Answer} \\\textbf{Length}} \\
		\cline{1-6}\cline{4-4}
		DROP~\cite{DBLP:conf/naacl/DuaWDSS019}
		&  EN     &  4.6K & \makecell{crowdsourcing}   & 2.5   & 2.0    \\
		Quoref~\cite{DBLP:conf/emnlp/DasigiLMSG19}
		&  EN     &  1.9K & \makecell{crowdsourcing}   & 2.5   & 1.6      \\
		MultiSpanQA~\cite{li2022multispanqa} 
		&  EN    &  6.5K & crowdsourcing   & 2.9   & 3.0  \\
		CLEAN (ours)
		&  CN    & 4.2K & real-world users   & 5.9   & 13.2   \\
		\bottomrule
	\end{tabular}
\end{table}

\begin{figure}[thb!]
	\centering
	\includegraphics[width=0.96\columnwidth]{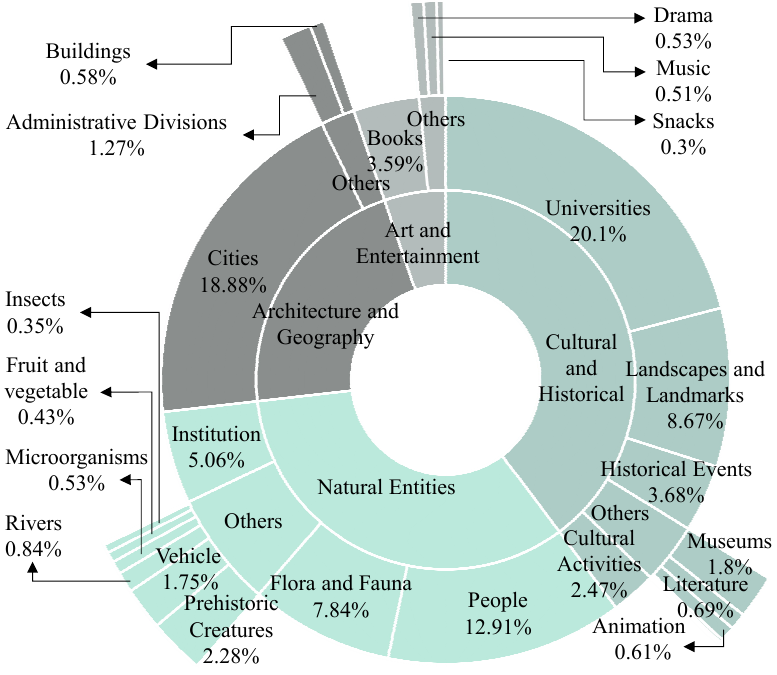}
	\caption{Subjects of questions in CLEAN.}
	\label{fig:subjects}
\end{figure}

CLEAN also has a broad coverage of open-domain topics. As shown in~\figref{fig:subjects}, the majority of questions in CLEAN are about \textit{cultural and historical}, \textit{architecture and geography}, \textit{art and entertainment}, etc.
To further examine and analyze the performance of different models on various question types, we categorize the samples into three types: \emph{description}, \emph{entity} and \emph{numeric}, based on the expected answer type. \tableref{tab:answer_types} illustrates the breakdown of these types along with an example for each answer type class. 
\begin{table}[!h]
	\small
	\caption{Proportion and examples of answer types in MultiSpanQA and CLEAN Datasets.}
	\label{tab:answer_types}
	\begin{tabular}{llcp{3cm}}
		\toprule
		\textbf{Dataset} & \textbf{Answer Type} & \textbf{\%} & \textbf{Example} \\
		\midrule
		\multirow{3}{*}{MultiSpanQA} & Description & 16.40 & \emph{other gases} \\
		& Entity & 76.30 & \emph{Vermont} \\
		& \emph{Numeric} & 7.30 & \emph{9,677 ft} \\
		\midrule
		\multirow{4}{*}{CLEAN} & Description & 76.54 & \emph{长满了尖锐的刺 (full of sharp thorns)} \\
		& Entity & 18.00 & \emph{北京市 (Beijing)}  \\
		& Numeric & 5.46 & \emph{1006万人~(10.06 million people)} \\
		\bottomrule
	\end{tabular}
\end{table}

\section{Experiments}
\label{sec:experiments} 

\subsection{Datasets}
\label{sec:datasets}
We conduct our experiments on two multi-answer RC datasets: CLEAN (our dataset) and MultiSpanQA~\citep{li2022multispanqa}. Dataset statistics are shown in~\tableref{tab:statistics} and~\tableref{tab:answer_types}. Note that we adopt the train/dev/test splits defined by Huang et al.~\citep{huang2023spans} for MultiSpanQA, as the official test set is not publicly available.
Next, we introduce the comparison models in our experiments. 

\subsection{Models}
\label{sec:baselines}
We explore 6 models drawn from related literature for multi-span answer extraction as baselines of CLEAN. 
Specifically, \textbf{SSE}~\citep{DBLP:conf/naacl/DevlinCLT19} is a single-span model that predicts the start and end positions of answer spans using a learnable linear layer on top of the encoder. It is adapted for multi-span question answering by treating the first span of a multi-span answer as ground truth during training.
More recently, \textbf{ITERATIVE}~\citep{DBLP:conf/acl/ZhangL0LF023} enhances this adaptation by iteratively appending  extracted answers to the question, inserting the word \textit{except} between them, and feeding  the updated question back into the single-span model until no more answers are identified. \textbf{TASE}~\citep{segal2020simple} employs a BIO tagging scheme to tag tokens and extract multi-span answers. \textbf{SpanQualifier}~\citep{huang2023spans} proposes a span-centric approach that learns enumerated span representations and extracts the most qualified spans as answers. 
In contrast, \textbf{LIQUID}~\citep{lee2023liquid} employs question generation as a form of data augmentation to enhance answer extraction performance, differing from the conventional emphasis on the exploration of model architectures.
Following Zhang et al.~\cite{DBLP:conf/acl/ZhangL0LF023}, we also develop a generative model for multi-answer extraction in CLEAN, utilizing the large language model \textbf{GPT-3.5} under the \textit{one-shot} setting. 
%

\begin{table*}[thb]
	\caption{The comparison results for all baselines, including EM F1 and PM F1. }
	\label{tab:alleval}
	\begin{tabularx}{0.98\textwidth}	{p{0.2cm}p{1.2cm}p{0.85cm}p{0.85cm}@{}cp{0.85cm}p{0.85cm}@{}cp{0.85cm}p{0.85cm}cp{0.85cm}p{0.85cm}@{}cp{0.85cm}p{0.85cm}@{}cp{0.85cm}}
		\toprule[0.3mm]
		\multirow{2}{*}{\rotatebox{90}{Dataset}} & \multicolumn{1}{c}{\multirow{2}{*}{\makecell[l]{Answer\\Type}}}  & \multicolumn{2}{c}{SSE}   &\multicolumn{1}{l}{} &
		\multicolumn{2}{c}{TASE}   & \multicolumn{1}{l}{} &
		\multicolumn{2}{c}{LIQUID} &\multicolumn{1}{l}{} &
		\multicolumn{2}{c}{SpanQualifier} & \multicolumn{1}{l}{} &
		\multicolumn{2}{c}{ITERATIVE}& \multicolumn{1}{l}{} &
		\multicolumn{1}{c}{GPT-3.5} \\ 
		\cline{3-4} \cline{6-7} \cline{9-10}  \cline{12-13} \cline{15-16}  \cline{18-18}  	\addlinespace 
		& \multicolumn{1}{c}{}  & \multicolumn{1}{c}{{\small{\makecell[l]{EM/PM\\(BERT)}}}} & \multicolumn{1}{c}{{\small{\makecell[l]{EM/PM\\(RoBE)}}}} & \multicolumn{1}{l}{} &\multicolumn{1}{c}{{\small{\makecell[l]{EM/PM\\(BERT)}}}} & \multicolumn{1}{c}{{\small{\makecell[l]{EM/PM\\(RoBE)}}}} &\multicolumn{1}{l}{} & \multicolumn{1}{c}{{\small{\makecell[l]{EM/PM\\(BERT)}}}} & \multicolumn{1}{c}{{\small{\makecell[l]{EM/PM\\(RoBE)}}}} &\multicolumn{1}{l}{} & \multicolumn{1}{c}{{\small{\makecell[l]{EM/PM\\(BERT)}}}} & \multicolumn{1}{c}{{\small{\makecell[l]{EM/PM\\(RoBE)}}}} &\multicolumn{1}{l}{} & \multicolumn{1}{c}{{\small{\makecell[l]{EM/PM\\(BERT)}}}} & \multicolumn{1}{c}{{\small{\makecell[l]{EM/PM\\(RoBE)}}}} &\multicolumn{1}{l}{} & \multicolumn{1}{c}{{\small{EM/PM}}}  \\ 
		\toprule[0.3mm]
		\multirow{4}{*}{\rotatebox{90}{CLEAN}}   & Descriptive   & 21.7/46.5 & 23.8/47.6   & & 33.6/70.5   & 39.3/73.5   & & 38.6/67.7  &44.6/73.5   &  &   51.9/74.5 & 52.9/76.6& &  34.3/66.0  & 34.2/68.9   & &   15.6/66.0   \\
		& Entity   & 33.2/53.6 &  39.0/55.9  & & 55.3/67.2   & 58.4/68.8   & & 53.4/76.0  &67.4/82.1   &  &   80.5/90.6 & 84.7/92.9& & 59.1/77.1   & 61.0/80.3   & &   30.0/64.9   \\
		& Numeric    & 57.1/74.7 &  53.1/76.8  & & 30.2/65.0  & 33.8/69.4   & & 53.3/74.9  &60.0/77.3   &  &   56.9/75.8 & 55.3/78.2& & 29.5/59.5   & 33.9/59.1   & &   14.5/71.2   \\
		& Overall    & 24.7/48.4 &  27.2/49.8  & & 39.1/69.2   & 44.4/72.0   & & 42.6/69.9  &50.6/75.7   &  &   60.4/79.2 & 62.0/81.3& & 38.5/68.3   & 38.8/70.6   & &   17.4/66.2   \\ 
		\cline{1-18} 
		\addlinespace
		\multirow{4}{*}{\rotatebox{90}{\small{MultiSpanQA}}}    & Descriptive   & 19.0/50.6 &  19.7/60.5  & & 33.6/65.1   & 46.1/74.2   & & 36.7/66.1  &41.6/72.3   &  &   43.2/62.2 & 53.6/72.7& & 32.8/58.7   & 43.5/70.2   & &   30.2/65.3  \\
		& Entity   & 27.4/57.6 &  29.4/59.2  & & 63.3/78.6   & 71.4/84.1   & & 67.2/81.7  &73.6/85.8   &  &   71.8/82.4 & 74.3/84.8 & & 67.4/78.8   & 73.5/84.8   & &   73.1/86.1   \\
		& Numeric    & 24.3/57.0 &  32.4/60.8  & & 29.2/58.1   & 40.6/67.6   & & 34.3/60.8  &41.4/69.3   &  &   45.3/65.4 & 53.2/72.3& & 36.9/55.6   & 51.0/72.4   & &   39.2/69.7    \\
		& Overall    & 26.3/56.8 &  28.3/59.5  & & 57.8/76.0   & 66.7/82.0   & & 60.5/79.2  &69.0/81.2   &  &   67.5/79.5 & 71.0/82.9& & 65.5/82.5   & 69.0/82.7   & &   65.5/82.5    \\ \addlinespace  \bottomrule
	\end{tabularx}
\end{table*}

\subsection{Experimental Setup}
\label{sec:setup}
\subsubsection*{Implementation Details} We utilize the HuggingFace implementation of $\text{BERT}_\text{base}$~\cite{kenton2019bert}\footnote{huggingface.co/bert-base-uncased}\footnote{huggingface.co/bert-base-chinese} or  $\text{RoBERTa}_\text{base}$\cite{liu2019roberta,cui-etal-2021-pretrain}\footnote{huggingface.co/FacebookAI/roberta-base}\footnote{huggingface.co/hfl/chinese-roberta-wwm-ext} as the \emph{encoder} with a maximum input length of $512$ tokens and a document stride of $128$ tokens to handle long passages. For training, we initialize the learning rate at $1e-4$, set a default batch size of $8$, and use the BERTAdam optimizer with a weight decay of 0.01. Exceptions to these settings are applied for models~\citep{lee2023liquid,DBLP:conf/acl/ZhangL0LF023,huang2023spans} with originally reported detailed configurations, and we follow them accordingly.
\subsubsection*{Metrics}
We adopt the official automatic metrics of MultiSpanQA~\cite{DBLP:conf/emnlp/DasigiLMSG19} for evaluation, including the precision(P), recall(R), and F1 in terms of exact match (EM) and partial match (PM). The EM shows the quality of predictions straightforwardly. However, counting the number of exact matches makes the score discrete and coarse. Besides, it penalizes the partial matched prediction too much. The PM score considers the overlap between the prediction and gold and thus can be used as a complementary metric to the EM metric.

\subsection{Experimental Results and Analysis}
\label{sec:evaluation}
We evaluate all baselines (\secref{sec:baselines}) on the test sets of two datasets (\secref{sec:datasets}) using both EM and PM F1 metrics (\secref{sec:setup}). 
The comparison results, including the overall performance and the performance across various instance types, are shown in \tableref{tab:alleval}.

We can see that all models demonstrate inferior overall performance when extracting answer spans from the CLEAN dataset compared to the MultiSpanQA dataset. This observation can be attributed to the higher prevalence of descriptive answer type instances in CLEAN.
These instances typically feature longer answers entwined with complex semantic relationships, making them more challenging to extract comprehensively using various model paradigms. 
As for SpanQualifier, the current state-of-the-art (SOTA) model for multi-answer RC, it achieves  EM F1 scores of 60.4\% using BERT and 62.0\% RoBERTa on the CLEAN dataset.  These scores lag behind those on MultiSpanQA by approximately 7\% for BERT and 8\% for RoBERTa in terms of EM F1. Moreover, SpanQualifier achieves only 51.9\%/74.5\% EM/PM F1 with BERT and 52.9\%/76.6\% EM/PM F1 with RoBERTa on descriptive type instances of the CLEAN dataset, which underscores the main challenge of answer extraction. 

On the other hand, the adaptation of single-span models for multi-answer extraction is less effective on the CLEAN dataset compared to MultiSpanQA. Specifically, the ITERATIVE model shows a notable decline in performance on CLEAN across various encoders. This might be result from the iterative use of the \textit{except} prompt, which is more adept at coordinating entity-type answers that are more common in MultiSpanQA. The CLEAN dataset contains answers with more complex semantic relations and presents a substantial challenge for current models.

Additionally, we observe that the performance of the one-shot GPT-3.5 model in multi-answer extraction falls short compared to smaller extractive models. 
GPT-3.5 tends to provide more detailed responses or rephrase  information  from the original context, even when prompted to avoid doing so. Such behavior is not ideal for tasks that require accurate and direct extraction of information, as it may introduce ambiguity or deviation from the source material.
\section{Related Work}
\label{sec:related}
Earlier extractive RC datasets, such as SQuAD~\citep{rajpurkar2016squad}, SQuAD2.0~\citep{rajpurkar2018know} and  HotpotQA~\citep{yang2018hotpotqa}, contain only single-span instances, where answers are limited to a single text span drawn from the provided context. These datasets are built by having humans read a given context, write questions and choose a specific answer span, the annotators may tend to make use of words from the answer text which potentially makes  their questions easier to answer. 
To mitigate this bias, TriviaQA~\citep{DBLP:conf/acl/JoshiCWZ17}, sourced from trivia enthusiasts, and QuAC~\citep{DBLP:conf/emnlp/ChoiHIYYCLZ18}, which restricts annotators from viewing the full context, are introduced. 

However, a comprehensive answer to a real-world question may consist of multiple text spans, or even the question itself could have multiple intents, where the answer to each intent is composed of one or more spans. Natural Question~\citep{DBLP:journals/tacl/KwiatkowskiPRCP19} and Quoref~\citep{DBLP:conf/emnlp/DasigiLMSG19} feature such multi-answer instances, sourced from Google search engine queries and designed for co-reference resolution, respectively. Despite their innovation, multi-answer instances are infrequent, at 2\% and 10\%. 
The DROP dataset~\citep{DBLP:conf/naacl/DuaWDSS019} presents complex questions requiring  discrete reasoning and arithmetic operations, with answer spans almost numeric and semantically homogeneous. 
While the MASH-QA dataset~\citep{DBLP:conf/aaai/PangLGXSC19} specializes in healthcare, expanding the answer space to general text types.
Recently, Li et al.~\cite{li2022multispanqa} propose the MultiSpanQA dataset, which consists of open-domain multi-span questions. The MultiSpanQA dataset is derived from Natural Question, whose questions are real queries issued to the Google search engine. 
Each question is associated with a context extracted from a retrieved Wikipedia page.  
MultiSpanQA also has an expanded variant by introducing single-span and unanswerable questions, namely MultiSpanQA (expand). 

The above datasets are all in English. 
CMQA ~\citep{DBLP:conf/coling/JuWZZ0022} is the first public multi-span QA dataset in Chinese, introducing a new annotate scheme that labels both fine-grained answers and conditions  as well as the hierarchical relations in between.
However, it formulates a new task of conditional question answering.
To ameliorate the situation of the lack of Chinese multi-span RC datasets, we propose a Chinese multi-span question answering dataset (CLEAN), which consists of multi-span questions in open domain and supports to cast RC as an answer extraction task.

\section{Conclusion}
\label{sec:conclusion}
In this paper, we propose  CLEAN, a comprehensive open-domain Chinese multi-span question answering dataset that includes a wide range of topics. The final release of CLEAN consists of 9,063 instances, with approximately 76\% being descriptive answer type of instances. Additionally, we have also established baselines from related literature to benchmark performance on CLEAN. Data analysis and experimental results show the main characteristics and challenges of CLEAN, indicating that CLEAN present a significant challenging for the community to solve.

\bibliography{refs}
%
%
%
%
%
%
%
%
%
%
\end{CJK*}
\end{document}